**Artificial Neural Networks that Learn to Satisfy Logic Constraints**

Gadi Pinkas* and Shimon Cohen

Afeka Tel-Aviv Academic College of Engineering and

* Gonda Brain Research Center, Bar Ilan University, Israel

**Abstract**

Logic-based problems such as planning, theorem proving, or puzzles like Soduko, typically involve combinatorial search and structured knowledge representation. Artificial neural networks (ANNs) are very successful statistical learners; however, for many years, they have been criticized for their weaknesses in representing and in processing complex structured knowledge which is crucial for combinatorial search and symbol manipulation. Two neural architectures are presented (Symmetric and RNN), which can encode structured relational knowledge in neural activation, and store bounded First Order Logic (FOL) constraints in connection weights. Both architectures learn to search for a solution that satisfies the constraints. Learning is done by unsupervised "practicing" on problem instances from the same domain, in a way that improves the network-solving speed. No teacher exists to provide answers for the problem instances of the training and test sets. However, the domain constraints are provided as prior knowledge to a loss function that measures the degree of constraint violations. Iterations of activation calculation and learning are executed until a solution that maximally satisfies the constraints emerges on the output units. As a test case, block-world planning problems are used to train networks that learn to plan in that domain, but the techniques proposed could be used more generally as in integrating prior symbolic knowledge with statistical learning.

*Keywords:* artificial neural networks, planning, constraint satisfaction, unsupervised learning, logic, neural-symbolic integration, binding problem

**Introduction**

The use of symbol processing and knowledge representation is fundamentally established in the field of Artificial Intelligence as a tool for modelling intelligent behaviour and thought. Specifically, logic-based systems have been used for years to model behaviour and cognitive processes. Nevertheless, it is believed that pure symbolic modelling is not enough to capture the adaptively, sub-symbolic computation, robustness, and parallelism of the kind demonstrated by neural networks in the brain (Kowalsky, 2011; Valiant, 2008).

ANNs are used today mainly for statistical pattern recognition, classification, and regression, and they have been criticized for their weakness in representing, manipulating, and learning complex structured knowledge (Fodor & Phylyshyn, 1988; McCarthy, 1988 ). For example, in areas such as vision and language processing, ANNs excel as classifiers and predictors. However, they are not very successful in recognizing and manipulating complex relationships among objects in a visual scene or in processing complex sentence structures governed by non-trivial semantic and syntactic constraints.

The main motivation of this paper is therefore to enable ANNs to have learnable combinatorial search and relational representation capabilities while using "practicing" to speed-up the search. Having such capabilities without sacrificing their statistical learning abilities will allow integration of the symbolic and sub-symbolic approaches for performing high-level cognitive tasks as motivated in previous research (Garcez, Lamb, & Gabbay, 2009; Hammer & Hitzler, 2007; Samsonovich, 2012; Valiant; 2008).

Several fundamental questions arise in the effort to represent and process complex knowledge using ANNS. First, there is a problem of representation and coding: How can objects and complex relational structures be encoded in the activation of neurons? This should be done without exponential explosion of network size (Feldman, 2013), and without losing accuracy as structural complexity grows (Plate, 2003). Then, there is the problem of encoding learnable relational knowledge in Long Term Memory (LTM), i.e., in connection weights. Also puzzling is the way to dynamically translate between the two forms of knowledge, i.e., retrieving structured relational knowledge from LTM into activation encoding and in the opposite direction, storing new and revised memories in the LTM for later use. Another problem is related to parallelism and to the network dynamics: How can a massive number of very simple, homogeneous units, organized in a relatively uniform way, collaborate asynchronously, yet still obtain globally-sound solutions without central control? Finally, how can learning be done without the presence of a teacher who knows the correct answer for the problem at hand?

The network architectures proposed, "Recurrent Network for Constraint Satisfaction" (CONSRNN) and "Symmetric Network for Constraint Satisfaction" (CONSyN), enable compact encoding of attributes and relationships in neural activation and storage of long-term relational constraints in connection weights. When provided with a problem to solve, such networks can execute iterative cycles of unit activation and learning, until a satisfying solution emerges on the output units. By unsupervised practicing on a training set composed of a few problem instances, the network can learn and improve its speed at solving unseen instances from a test set.

The two neural architectures are described, theoretically analysed, and empirically tested. In the implementations presented, high-order multiplicative connections (sigma-pi units) are used with standard activation functions: sigmoidal in CONSRNN and binary-threshold in CONSyN.

The proposed ANNs facilitate learning by minimizing a loss function that is based on the domain constraints, which are provided a priori (prior knowledge). When the loss is minimized, it means that the output activation values encode a solution that minimizes the constraint violation.

The paper is organized as follows: Chapter 2 illustrates an example of a simple block-world planning problem that will be learned and searched by the proposed ANNs.. In Chapter 3, we show how to represent the inputs and the solution output of the planning problem in activation of the network's visible units. More generally, the principles for a general-purpose, expressive yet compact binding mechanism are provided. Chapter 4 illustrates two different neural architectures for constraint satisfaction: CONSRNN, a network based on simple recurrence loop, and CONSyN, based on symmetric connections. Thereafter, sigma-pi units are reviewed, and two types of loss functions are introduced which measure how well a solution satisfies the constraints of the problem. The detailed activation-learning algorithm of CONSyN and its mathematical derivation are described in Chapter 5, while Chapter 6 describes the algorithm developed for CONSRNN. Chapter 7 describes the experimental framework and provides experimental results from the block-world planning domain. Chapter 8 discusses related disciplines, offers a conclusions, and suggests directions for future research.

## A Simple Planning Problem and Its Constraints

Although, this article is about a general technique to learn logic constraints, for illustration, we have chosen the well known problem of planning in block-world with a simple intuitive reduction to FOL. Consider a world of blocks with properties such as *color* and *size*. Some blocks happen to be on the floor, while other blocks are arranged "above" others. A planning agent has the task of arranging the blocks by a series of moves, so that a certain "goal" configuration of blocks is obtained. The block-world changes with time, as the agent moves "cleared" blocks (i.e., those with no other block above them) and puts them either on the floor or on top of other blocks. The result of the planning process is a "solution plan" which details a series of valid block moves and corresponding block configurations. A solution plan starts with the initial block configuration at time $t_0$ and lists the moves that need to be made at each time step ($t_1, t_2…$) until the goal configuration is obtained at time $t_{Final}$.

The relation *Above(i,j,t)* specifies that block *i* is above block *j* at time step *t*. *Color (i,"R")* designates that block *i* is red, while *Size(i,"L")* designates that block *i* is large. Note that for ease of problem specification, few other relations are used:

    *Move(i,t)*	// block *i* moves at time *t*.
    *Cleared(i,t)*	// block *i* at time *t* is cleared (nothing on top).
    *Floor(i)*	// object *i* is a floor

A planning problem instance is specified by an initial configuration of blocks and a goal configuration, like the example in Figure 1: The initial configuration includes four blocks arranged as in Figure 1. The goal configuration is to build a modified tower as in Figure 1b.

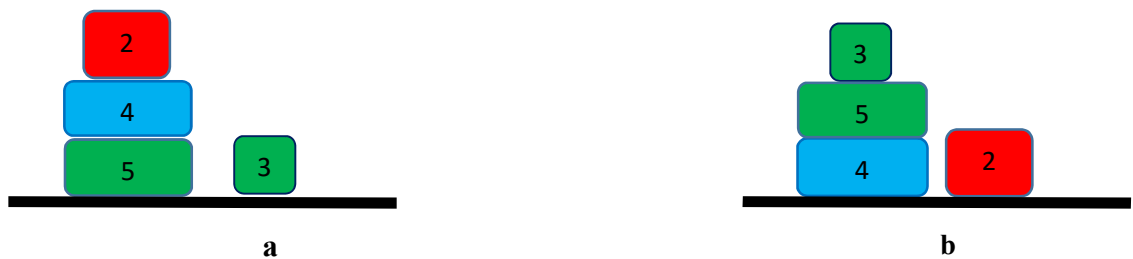

*Figure 1.* A planning problem instance: a) initial block configuration, b) goal configuration

The plan that the agent must generate is a valid series of up to *k* moves, which gradually changes the initial configuration until the goal configuration is reached. A plan is valid if the logic constraints of Table 1 hold.

Table 1: Constraints of a simple block-world planning problem.

| | |
|---|---|
| ($\forall i,j,k,t>0$) Above(i,k,t) → Above(i,k,t-1) ∨ Move(i,t-1)) | If one object is above another at time t, it must be above that object in the previous step or have been moved in the previous step. |
| ($\forall i \neq j,k,t$) Above(i,k,t) ∧ Above(j,k,t) → Floor(k) | If two objects are above an object then that one object is the floor. |
| ($\forall i,t$) ¬Above(i,i,t) | No object can be above itself. |
| ($\forall i,t,k \neq i$) Above(i,j,t) →¬Above(i,k,t) | An object cannot be above more than one object. |
| ($\forall i,j,t$) Floor(j) →¬Above(j,i)<br>($\forall i,j,t$) ¬Floor(j) ∧ Above (i,j,t) → ¬Cleared(j,t) | The floor cannot be above any object:<br>If an object is above a second object that is not floor, then the second is not cleared. |
| ($\forall i,t$) Move(i,t<K) → Cleared(i,t) ∧ Cleared(i,t+1) | If an object is moved at time *t*, then it is cleared at time *t* and at time *t+1*. |
| ($\forall i,j,t$) Move(i,t<K) ∧ Above(i,j,t+1) → Cleared(j,t) | If an object was moved at time *t* to be above a 2nd block, then the 2nd was cleared at time *t*. |
| ($\forall i,t, \exists j$) Move(i,t) → Above(i,j,t) | If an object was moved, then it was above something. |
| ($\forall j,t$) Cleared(j,t) ∧ (($\forall i$) ¬Above(i,j,t)) → Cleared(j,t+1) | If an object is cleared at time t and has nothing above it, it will stay cleared at t+1 (frame axiom) |
| ($\forall i,s \neq s'$) Size(i,s) → ¬Size(i,s')<br>($\forall i,c \neq c'$) Color(i,c) → ¬Color(i,c') | An object must have only one size and color. |

Every valid plan must satisfy the above constraints, yet in order to produce parsimonious plans, soft constraints may also be added, discouraging unnecessary moves:

$$(\forall i,t) \neg(Move(i,t))$$

Similarly, other soft constraints such as $(\forall i,j,t) \neg(Above(i,j,t))$ may also be added to avoid unnecessary "Above" relationships.

Note that the frame constraints specifying that an object stays cleared unless something is above it are not really necessary. Eliminating the frame constraints while satisfying the other constraints still generates valid plans in the *Above* crossbar (while the *Cleared* crossbar interpretation turns a bit counterintuitive).

In a more general formulation, each constraint may be augmented by a positive number, called penalty in Pinkas (1994), weight in Markov Logic networks (Domingos 2008), or confidence in Tran and Garcez (2016) and Penning et al. (2015). In our simple block-world planning domain, a penalty of *1000* is used to specify hard constraints, and a penalty of *1* is used for soft constraints.

In block-world planning of *bounded length*, the size of the maximal plan is restricted ($K$) and so is the maximal number of objects ($N$). These bounds are necessary, as the full solution output should be encoded in using a finite set of network units. These bounds are also used for converting the above FOL expressions into Conjunctive Normal Form (CNF), by replicating the constraints according to the indices specified. Thus for example, $(\forall i,t, \exists j)\ Move(j,t) \rightarrow Above(i,j,t)$ is translated into a CNF by replicating:

for i=1 to N
  for t=1 to K
    AssertClause ("¬Move(i,t) ∨ Above(i,1,t) Above(i,2,t) …∨ Above(i,N,t)")

All planning instances including those size (t) and object cardinality (n) smaller that the bounds (n≤N and t≤K), share the same propositions and the same hard and soft constraints specifying what a valid plan is, yet each planning instance is different in its initial and goal states. These initial and goal states are considered the inputs of the planning problem and can also be stated as simple conjunctive constraints. For example, the following conjunctions specify the initial configuration of Figure 1(a):

$Color(2,Red) \wedge Color(4,Blue) \wedge Color(5,Green) \wedge Color(3,Green) \wedge Floor(1) \wedge Above(2,4,t_0) \wedge Above(4,5,t_0) \wedge Above(5,1,t_0) \wedge Above(5,1,t_0) \wedge Above(3,1,t_0)$

where $t_0$ is the first time step. The goal configuration at the final step of the plan, as in Figure 1(b), is specified by:

$Above(3,5,t_K) \wedge Above(5,4, t_K) \wedge Above(4,1, t_K\ 7) \wedge Above(2,1, t_K)$,

where $t_K$ is the final state.

A valid solution for the planning instance of Figure 1 is illustrated in Figure 2. It consists of a series of moves in time:

$Move(2,t_0), Move(4,t_1), Move(5,t_2), Move(3,t_3)$

and corresponding block configurations at each time step, which are consistent with all the hard constraints specified. In Figure 2 at time $t_0$, the initial block configuration holds, and only Blocks 2 and 3 are cleared. By moving Block 2 onto the floor, the configuration at $t_1$ is created, where all blocks are cleared except block 5. At $t_1$, Block 4 is moved onto the floor. At $t_2$, the large green

Block 5 is moved, and at *t3* Block 3 is moved to create the desired goal configuration at time *t₄*. No block moves further, so that the same block configuration remains static until the last time slot. The configuration of the last step therefore satisfies the goal constraints.

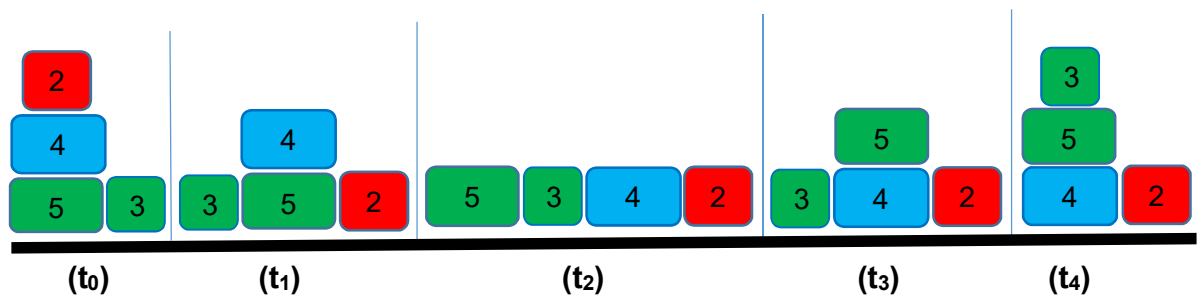

*Figure 2.* The solution plan to the planning instance of Figure 1

Randomly generated planning instances are used for training and testing. Each problem instance differs from others only in the initial and goal configurations; the rest of the constraints (enforcing plan validity) are shared among the different instances.

A single ANN is created for solving any such planning instance of up to $N$ blocks and up to $K$ steps. This architecture is independent of the specific constraints that will be learnt and be stored as weights. The generated ANN will start solving a planning instance problem by first clamping the Boolean values of the Initial-Goal input onto the visible units of the network. In the following chapter, the structure of the visible units of such ANN is described.

## Representing Relational Knowledge in Unit Activation

To attach attributes to objects and to represent relationships among them, an expressive yet compact mechanism for binding objects, attributes, and variables is needed. In the following, such a binding mechanism is illustrated that can encode block-world objects, their attributes (e.g., *red*, *small*, *cleared*), and their relationships with other objects.

**Forming Bindings between Block Objects and their Properties**

In the block-world planning problem of the figures above, the number of block objects is bound to N=5 and the number of maximal planning steps to K=6. Blocks can be of three different colors and three different sizes.

To provide a "glue" that will allow the binding of objects with their properties, collections of units called binders (*b1…b5*) are allocated to each of the objects that might participate in the plan. The network visible units consist of a pool of such binders. When there is a need to bind an object with some attributes, one of the binders in the pool must be allocated for the object and should refer to the appropriate attribute values. For example, binder *b3* is a collection of units that was allocated to the "small green block" *b5* to the "large blue block," and *b1* was allocated to represent the floor. Each such binder is capable of encoding references to attribute values (e.g. *small*, *green*) and to other object binders thus forming relationships between objects. The binders are organized in 2D and 3D matrices of units that are called crossbars. Crossbars are matrices of units, each with an activation function that can be interpreted as a Boolean value. Using multiple crossbars, a binder may encode several properties and relationships.

Figure 3 shows two such crossbars representing the binding of colors to 5-block objects and similarly the binding of sizes to the same objects. Binder 2 (*b2*) consists of the 2$^{nd}$ rows of the *color* and *size* crossbars and is allocated to a medium green block. Binder 3 (*b3*) is allocated to a small green block, and Binder 5 (*b5*) is a large red block. The first row (*b1*) is allocated to the floor. A cell marked with a 1 represents a firing unit (*Boolean true*). An empty cell represents a non-firing unit (*false*). Because in the planning domain, each object has only one color and only one size (i.e., many-to-one relationship), the units in each row of a crossbar are mutually exclusive. These kinds of domain constraints are added to the constraints described in the previous chapter and will lead to a more compact representation provided later in this chapter.

| Color | Red | Green | Blue |
|---|---|---|---|
| b1 |  |  |  |
| b2 | 1 |  |  |
| b3 |  | 1 |  |
| b4 |  |  | 1 |
| b5 |  | 1 |  |

| Size | Small | Medium | Large |
|---|---|---|---|
| b1 |  |  | 1 |
| b2 |  | 1 |  |
| b3 | 1 |  |  |
| b4 |  |  | 1 |
| b5 |  |  | 1 |

*Figure 3.* Crossbars representing *color* and *size* properties bound to objects. Each row is a binder that ties an object with its properties.

**Encoding Relations**

The rows in a crossbar may also reference other binders, thus forming relationships among objects. In Figure 4, a crossbar, *Above(N,N)*, representing the binary relation, *Above(i, j)*, is illustrated. The crossbar encodes the fact that the object represented by Binder *b4* (large blue block) is above the object represented by Binder *b5* (the large green block). At the same time, *b5* is above *b1* (the floor), and two other objects (*b2*, *b3*) are also directly above the floor.

| Above | b1 | b2 | b3 | b4 | b5 |
|---|---|---|---|---|---|
| b1 |  |  |  |  |  |
| b2 | 1 |  |  |  |  |
| b3 | 1 |  |  |  |  |
| b4 |  |  |  |  | 1 |
| b5 | 1 |  |  |  |  |

*Figure 4:* A crossbar representing the relation *Above* using binders referencing other binders.

The unit activations in the crossbar of Figure 4 encode the *Above* relation at a certain time step: *Above(b2,b1), Above(b3,b1), Above(b5,b1), Above(b4,b5).* Thus, when the crossbars of Figure 3 and Figure 4 are combined, the activation of their units encode a description of the visual scene of Figure 2 at time ($t_1$).

Note that the *Above ()* relation, in this simple planning example, is a one-to-one relationship: a block cannot be above two blocks and no two blocks can be above a single block. It is interesting to note that after these constraints are learned in the proposed symmetric architecture, the rows and columns of such crossbars turn to be "winner-takes-all" units, similar to the wiring suggested by Hopfield and Tank (1985).

**Using 3D Crossbars.** For our planning example, the *Above* relationship should be 3D to capture the fact that the object configurations change over time. Therefore, the *Above(N,N,K)* crossbar is used, adding a time dimension, bounded by *K*. Thus, *Above(i,j,t)* means that object *i* is directly above object *j* at time *t*. Figure 5 illustrates the *3D Above* crossbar encoding the changes in the *Above* relation over time in the solution plan of Figure 2, starting at the initial configuration and ending at the goal configuration. The input to the planner consists of the initial and goal configurations (in bold), which are clamped on the *Above* crossbar at $t_0$ and $t_5$, respectively. The configurations encoded at time steps $t_1$ to $t_4$ are generated by the planning agent and represent the solution found. In the example, at time $t_1$, the *Above* crossbar encodes the configuration after moving the block of Binder 2 onto the floor. At $t_2$, as a result of moving Block 5, all blocks are directly above the floor and all are cleared. The goal is reached at $t_4$ after moving Block 3 above Block 5. At time $t_4$ no move is made, so the configuration remains static at $t_5$.

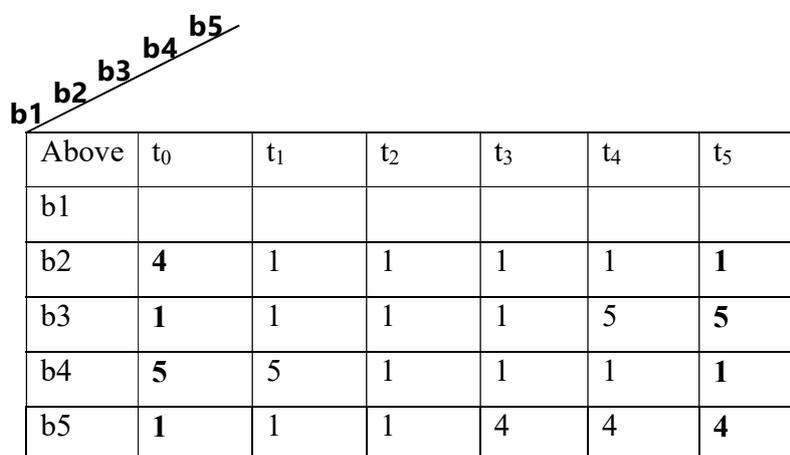

| Above | $t_0$ | $t_1$ | $t_2$ | $t_3$ | $t_4$ | $t_5$ |
|---|---|---|---|---|---|---|
| b1 | | | | | | |
| b2 | **4** | 1 | 1 | 1 | 1 | **1** |
| b3 | **1** | 1 | 1 | 1 | 5 | **5** |
| b4 | **5** | 5 | 1 | 1 | 1 | **1** |
| b5 | **1** | 1 | 1 | 4 | 4 | **4** |

*Figure 5:* The *Above* crossbar encodes the relationships between objects at each time step. The value *j* in cell *Above*(i,k) means: *Above*(i,j,k)=True. The bold numbers represent clamped inputs.

In Figure 6, the moves that should be executed by the plan are reflected within the crossbar, *Move(N,K)*; the block referenced by Binder 2 is moved at $t_0$, creating the configuration encoded by *Above(,,$t_1$)*. The block referenced vy binder 4 is moved at $t_1$, creating the configuration of *Above(,,$t_2$)*. Block 5 is moved at $t_2$ and finally, Block 3 is moved at $t_3$, generating the desired goal at *Above(,,t_5)*. The *Move* crossbar is not mandatory for representing the plan, as it can be deduced from the *Above(i,j,t)* crossbar, yet its existence helps to specify shorter and more intuitive constraints.

Similarly, the crossbars *Clear(N,K)* and *Floor(N)* are not mandatory but are also added for convenience. The *Clear* crossbar describes which blocks are cleared at the various time steps, while the *Floor* crossbar specifies which binder represents the floor object.

| Move | $t_0$ | $t_1$ | $t_2$ | $t_3$ | $t_4$ |
|------|-------|-------|-------|-------|-------|
| b1   |       |       |       |       |       |
| b2   | **1** |       |       |       |       |
| b3   |       |       |       | 1     |       |
| b4   |       | 1     |       |       |       |
| b5   |       |       | 1     |       |       |

*Figure 6.* The *Move(N,K)* relation encodes which objects are moved at each time.

The visible units of the planning ANN are made of the set of crossbars: *Color, Size, Above, Clear,* and *Floor*. These visible units consist of input units, which are clamped per planning instance according to the desired initial-goal configurations, and the output units, which encode the solution plan to be generated by the ANN. Thus, the clamped inputs are the units of *Above*, *Clear* and *Floor* crossbars at $t_0$ (for the initial configuration) and at $t_5$ (for the goal configuration), with *Color,* and *Size* for specifying the objects' attributes. The output units consist of crossbar units *Move* and *Above(,,t)*, where $K > t > 0$.

Note also that a binders-crossbar is not a simple adjacency matrix that characterizes the relation: the binders are pointers that indirectly reference to objects and may be allocated upon need from a binder pool. In a more realistic system, allocation constraints such as: *(∃i) Color(i,"Blue") ∧ Size(i,"Small")* could be used to allocate a binder when a small blue block is detected by the agent or when a goal involving object is provided . Thus, as in the binding mechanism suggested in (Pinkas, Lima, Cohen 2013) the number of binders needed could be much smaller than the number of objects in the domain.

**Using Compact Distributed Representations**

Relationships that are many-to-one may be represented compactly by using *N log(K)* units. For example, if each object in the domain can only have one of *K=7* colors and *N* is the number of objects supported, then the seven colors can be encoded using only three units and the size of the crossbar is only *3N*. Similarly, as each block can only be placed above one other block, each of the blocks below can be encoded using only *log(N)* units. If *K* is the maximal plan length, the size of the 3D *Above* crossbar is therefore *N·log(N)·K* units.

Theoretically, this compact representation reduces the search space. However, it may generate a biased input representation and may necessitate higher-order connections or deeper networks (as the complexity of the constraints grows). Due to its complexity, this compact form of representation was not tested experimentally.

The binder-pool approach can be extended to a general representation of any set of FOL formulae using just two generic crossbars. In Pinkas, Lima, and Cohen (2013), a general purpose binding mechanism is proposed based on crossbars that is able to represent any directed labeled graph. It uses a 2D *Symbol* crossbar that binds symbols to graph nodes and a 3D *Arc* crossbar that binds nodes to other nodes and labels these arcs. This general mechanism can encode any knowledge-base of FOL formulae in unit activation, while constraints are provided to allow processing of such formulae, such as retrieving them from LTM, unifying them, and generating bounded deductions. The size complexity of such network is O($N \cdot log(N) \cdot log(K)$), where $K$ is the knowledge base length and $N$ is the solution length.

## ANN Architectures for Constraint Satisfaction

In this chapter, two different types of ANN architectures are sketched for solving constraint satisfaction problems. The first uses a symmetric matrix of weights and is based on the energy minimization paradigm as in Hopfield's (1982) networks, Deterministic Mean-Field Theory (MFT) (Peterson & Anderson, 1987), Boltzmann (BM) Machines (Ackley, Hinton & Sejnowski, 1985), Belief Networks (Hinton, Osindero & Teh, 2006) and Restricted (RBM) BMs (Hinton & Salakhutdinov, 2006). The second is a based on Recurrent Neural Networks (RNN).

Both architectures have almost identical sets of visible units, where input values are clamped and where the output solution emerges as a satisfying solution. The rest of the units are hidden units, which, as shown later, can be traded with high-order connections. The visible units are directly mapped onto the problem's Boolean variables and should maximally satisfy the problem constraints. In the block-world example, the visible units are the units of the crossbars described earlier; the input units are those clamped by the initial and goal configurations, and the output units are those that at the end of the process encode the generated plan.

In order to solve a specific problem instance clamped on the input units, both network architectures iterate between activation calculation and learning phases. Activation calculation computes the activation function of each of the units, while learning changes the connection weights, trying to minimize a loss function that relates to constraint violations. The iteration stops when a measure of the violation is "small enough," meaning that the process continues until all the hard constraints are satisfied and the number of violated soft constraints is less than a given parameter.

The idea is that practicing on solving training problem instances from the same domain will speed-up the solving of unseen instances; i.e., while solving several problem instances from the same domain, weights are learned that "better" enforce the domain constraints, thus decreasing the number of activate-learn iterations needed to solve test instances from the same domain.

Figures 7 and 8 illustrate the two architectures. Despite their similarities, the two architectures are very different in their dynamics and in their learning procedures.

**Energy Minimization Symmetric Architecture**

Figure 7 illustrates a symmetric ANN architecture for constraint satisfaction (CONSyN). The visible units consist of the problem crossbars and allow the encoding of both the input and the output. The input (init-goal) configuration is clamped onto the input part of the visible units while at a fixed point; the non-clamped (output) visible units get activations that ideally satisfy the constraints. Symmetric (possibly high-order) weighted connections and (optional) hidden units enforce the problem constraints on the visible units. The target of learning is to learn such connection weights that effectively enforce the constraints.

Whenever, the network reaches a unit-configuration that is a stable fixed point (i.e., energy minimum), a check is made to determine whether constraints are still violated. If that happens, a learning step is made, and the weights are updated in a way that increases the energy of the unwanted unit configuration.

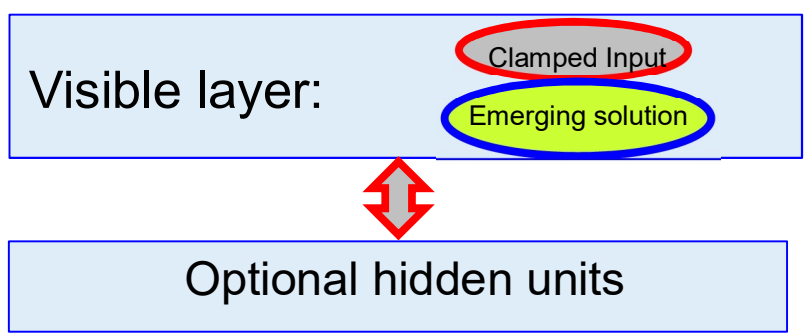

*Figure 7:* CONSyN architecture. The visible units are capable of encoding both the input and the desired solution. The hidden units are optional and may be traded with high-order connections.

*Algorithms 1:* Principles of activate-learn iterations in CONSyN architecture

Given a set of constraints, a problem instance input and a CONSy network:

a.  Clamp the input onto the input units (e.g. initial and goal configuration).
b.  Set random initial activation values to all non-clamped units.
c.  Until a fixed point is reached, calculate activations of the non-clamped units.
d.  While violation is not "small" enough do:
    i.  CONSyN learning (Algorithm 3).
    ii. Until a fixed point is reached, calculate activations of the non-clamped units (Algorithm 4).

Algorithm 1 describes in high-level the iterative procedure of activation calculation and learning until a "good enough" solution is found. When the activate-learn loop ends, the visible units have a violation measure that is "small enough" and at least satisfies all the hard constraints.

The learning in the symmetric case (see next chapter) may be viewed as increasing the "importance" of the violated constraints. Such change in constraint importance is translated to Hebbian connection weight changes that in turn increase the energy of that violating state. Thus, learning is actually re-shaping of the energy function by incrementally lifting the energy of violating local minima. The result of the training is a network with an energy function that resembles a violation loss function.

**Recurrent Network Architecture (CONSRNN)**

The RNN in Figure 8 uses directed connections (instead of symmetric in CONSyN). The input layer consists of crossbar units capable of encoding both the clamped problem input (e.g. initial-goal configurations) and activations copied from a feedback layer in previous recurrence. The output layer consists of crossbar units capable of encoding the generated plan. In our implementation, a simple feed-back loop connects the output units to the non-clamped input units. However, the architecture is not limited to this form of recurrence and future implementations could feed-back hidden units as-well.

Layers of hidden units, which could be traded with high-order connections, map the input layer into the output layer with the purpose of reducing constraint violations on the output units. After clamping the inputs, the activation of the output layer is calculated by forward propagation. The output layer units together with the clamped input units are taken as a possible solution and are checked for constraint satisfaction. If the violation of the output units is not "small enough," learning is done by backpropagation, and the output units are copied back into the non-clamped inputs (possibly with added noise) for another iteration of forward activation and learning. The clamped inputs stay clamped as in the previous iteration, and the process continues until a "good enough" solution emerges on the output units. Using a loss function that measures constraint violation, the network actually learns to greedily map the input into a solution that minimizes the violation, while the noise added within the copied activations helps in escaping from a local minimum of the violation function.

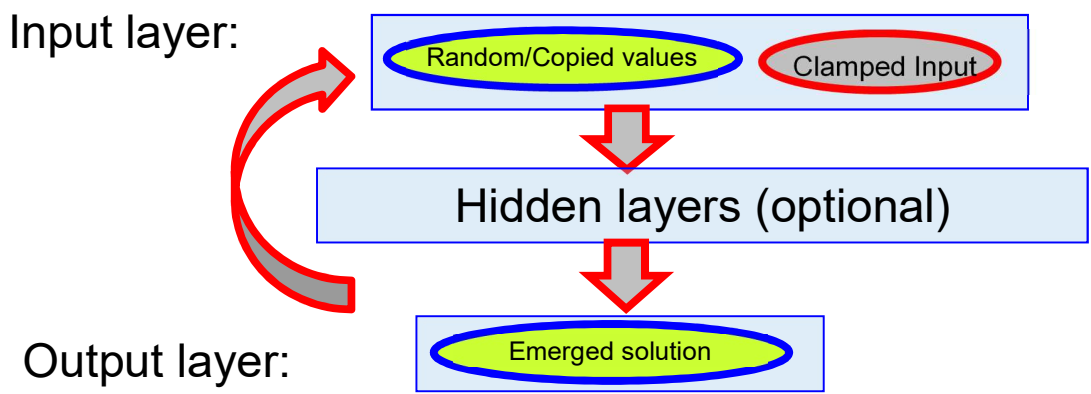

*Figure 8*: The CONSRNN architecture consists of a feedforward network with simple feed-back loop. Activations from the feedback layer are copied onto the non-clamped input layer.

*Algorithm 2:* High-level principles of activate-learn iterations in CONSRNN
   Given a set of constraints, a problem instance input values and a CONSRNN network:

   a. Clamp the input values on the input units.
   b. Set random initial activation values to all non-clamped input units.
   c. Compute the activation of the output layer (by performing feed-forward calculation).
   d. While the violation (output units and clamped inputs) is not "small enough," do:

i. Learning by back propagating the gradient of a violation loss function.
ii. Copy the feedback layer onto the (non-clamped) input layer.
iii. Add noise to the non-clamped input units.
iv. Compute the activation of the output layer.

As in the symmetric architecture, the search involves iterations of activation, violation check and learning. Algorithm 2 describes in high-level the iterative search which is terminated when a "good enough" solution is found (see Algorithm 5 for detailed implementation)

**Sigma-Pi Units with High-Order Connections and their Tradeoff with Hidden Units**

Unlike the classic units and pairwise synapses that are mainstream for many current neural models, the output of a sigma-pi unit is based on the sum of weighted contributions from multiplicative subsets of input values. Units of this type were proposed by Feldman and Ballard (1982) and have been used occasionally by other connectionist modelers, most commonly as gating devices allowing one unit to control the interconnection between other units (Sejnowski, 1986; Zhang, 2012).

A sigma-pi unit is a generalization of a "classic" unit that uses high-order connections in addition to biases and pairwise connections. A *k*-order connection connects a set *S* of *k* units using a single weight. It can be a directed high-order connection (as in the feed-forward network of Figure 9) or a symmetric connection (Figure 10). Sigma-pi units calculate the weighted sum of products of the input units as in Equation 1.

*Equation 1*

$$z_i = \sum_{S_{j,i}} w_{S_j} \prod_{k \epsilon S_{j,i} - \{i\}} X_k \tag{1}$$

To calculate the activation value for unit *i*, the weighted sum ($z_i$,) of the connection products directing to *i* is computed. After calculating $z_i$, unit *i* calculates its activation value $y_i$ using an activation function **σ** as in Equation 2.

*Equation 2.*

$$y_i = \sigma(z_i) \tag{2}$$

Although a variety of activation functions could be used, the binary threshold activation function was used in our implementation of CONSyN, whereas sigmoidal activation was used in CONSRNN.

In Figure 9, an example of a high-order feedforward network for XOR is shown with a 3[rd] order directed connection $\{A,B\}_C$ that connects the product of Units A and B with Unit C and uses a single weight of -4. Unit C also has a bias of -1 (1[st] order connection) and 2 pair-wise connections $\{A\}_C$, $\{B\}_C$ (2[nd] order) with weights of -2. Unit C therefore is a sigma-pi unit which calculates the weighted sum of input products: $Z_C = -4AB + 2A + 2B - 1$:

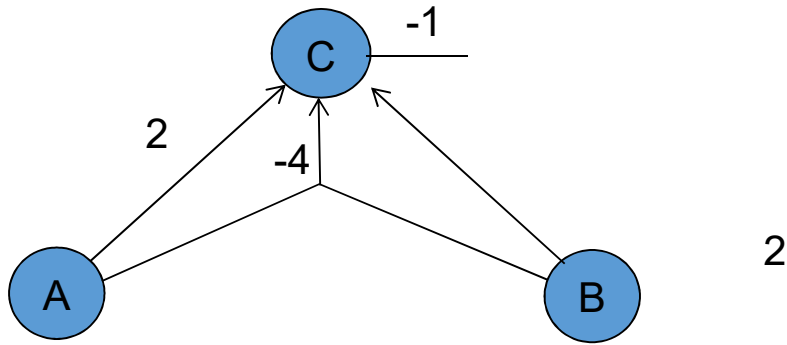

*Figure 9:* High-order feed-forward network for XOR: Unit B is a sigma-pi unit with a bias, 2 pairwise connections and a 3rd-order connection.

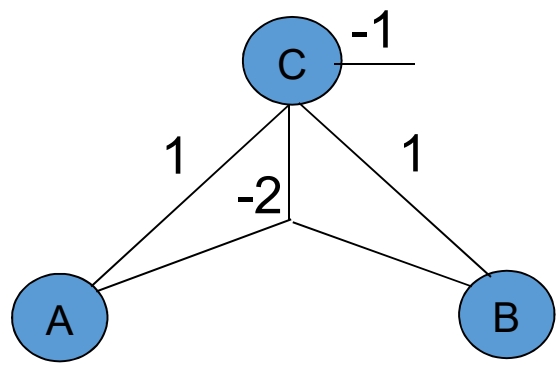

*Figure 10:* A symmetric 3rd order network with three sigma-pi units and a single 3rd order connection, searching to satisfy $C \rightarrow (A\ XOR\ B)$.

In the symmetric architecture, a $k$-order connection $S_i$ is treated as if there were $k$ such connections directing to each member of $S_i$, all with the same weight. Figure 10, illustrates an example of such high-order symmetric network. The network consists of four symmetric connections: A single 3rd order connection that connects all three units with weight of -2; two standard pairwise connections that connect units A with B and B with C with weight of 1 and a negative bias for Unit B. Unit A computes therefore $Z_A=C-BC$, Unit B computes $Z_B=-AC+C$, and Unit C computes $Z_C=-AB+A+B-1$. The network happens to search for a satisfying solution to *C→(A XOR B) as when C=1,* either *A* or *B* must be 1 exclusively.

Standard (pairwise) symmetric networks may be viewed as searching for a minimum of a quadratic energy function that is directly related to the weighted connections and may be written as a weighted sum of up to two variable products. Similarly, high-order symmetric networks minimize higher-order energy functions that sum the weighted products of the units in each of the connection subsets. Thus, the network shown in Figure 10 minimizes the 3$^{rd}$ order energy function: *E(A,B,C)= C-AC-BC+ABC*.

The connections are symmetric in the sense that a *k*-order connection is an undirected hyper-arc with a single weight, i.e. the tensor of weights is invariant under all permutations of its unit arguments. For example, in Figure 10, a single weight is attached to all connection permutations: $w_{\{A,B\}C}=w_{\{A,C\}B}=w_{\{B,A\}C}=w_{\{B,C\}A}=w_{\{C,A\}B}=w_{\{C,B\}A}=$ *-1*.

High-order connections can be traded with hidden units. Trivially so, in feedforward networks, a directed *k*-order connection that adds $w \prod_{k \in S} X_k$ to the sum of weighted inputs *z*, can be replaced by a single hidden unit that performs an *AND* on the inputs using standard pairwise connections and a bias. Similarly (but not as trivially), symmetric *k*-order connections in the energy minimization paradigm can be replaced by at most *k* units (Pinkas, 1991) with an equivalent energy function.

In the implementations of the two architectures described in this paper, high-order connections are used instead of hidden units. The lack of hidden units simplifies the implementation and seems to reduce experimentation time. On the other hand, high-order connections may easily cause overfitting, and this may explain the degradation in network performance which was occasionally observed. Nevertheless, future implementations may choose to trade the high-order connections with deep layers of hidden units and thus will not need sigma-pi units.

**Loss Functions**

ANNs typically use loss functions to measure their functional error. The gradient of this loss function is used to incrementally update the connection weights of the network so that gradually, the error is minimized. Both suggested architectures use loss functions that measure the degree of constraint violation of activation values of the visible units. Therefore, minimizing the loss is equivalent to minimizing constraint violations. In both implementations, learning is done using Stochastic Gradient Descent (SGD) by computing the loss gradient and performing a weight update step in the opposite direction.

To calculate the loss, we assume that the domain constraints are provided as a CNF. A CNF is a conjunction of clauses, and each clause is a disjunction of literals. Each literal is a negative or positive Boolean variable which in turn corresponds to a visible unit. The CNF violation is measured therefore with respect to the array of activation values of the visible units. A clause is satisfied if at least one of its positive literals refers to a unit activation that is interpreted as "True" or at least one of its negative literals refers to a unit with "False" activation. The Violation Loss function (*Vloss*) is a non-negative function that measures how "far" the activation values are from perfectly satisfying all the CNF clauses. A zero *Vloss* means that all clauses are perfectly satisfied, and a positive *Vloss* means that some clauses are violated, at least to a degree. More generally, in order to accommodate soft and hard constraints, as well as other probabilistic and non-monotonic logic systems (Pinkas, 1995; Domingos, 2008), each of the clauses in the CNF is augmented by a positive penalty ($\alpha_c$), which specifies the "strength" (or "confidence") of the constraint. The *violation* of such an augmented CNF is the sum of the penalties of the violated clauses.

The function *ClauseLoss(c,y)*, described in the following sections, measures the degree of violation of a *single* clause *c* with respect to an activations array *y*. The function returns zero when clause *c* is satisfied by *y* and returns a positive real number if *c* is violated to "some degree."

*Equation 3.*

$$Vloss(CNF, y) = \frac{1}{\sum_{c \in CNF} \alpha_c} \sum_{c \in CCNF} \alpha_c \, ClausLoss(c, y) \tag{3}$$

Equation 3 provides the *Vloss* function for measuring the violation of a given augmented *CNF*. It is the weighted average of the *ClauseLoss* functions for each clause in the CNF. From Equation 3, it follows that the gradient of the total *Vloss* is the average of the gradients per clause. In the following, two different formulations of the *ClauseLoss* function are provided: *Proximity Product* and *LogSat*.

**Proximity Product *ClauseLoss.*** The *ProP ClauseLoss* of a clause *c* with respect to activation array *y* is the product of the distances of the actual values $y_l$ of the literals *l* in the clause from the desired values of that literals (see Equation 4). The *ProP ClauseLoss* function outputs a real number in the range *[0,1]*, where 1 means perfect violation, and 0 means perfect satisfaction.

*Equation 4.*

$$ProP(c, y) = \prod proximity(l, c, y_l) \tag{4}$$

The *proximity* function of Equation 5 measures the distance between the actual activation of unit *v* and its desired literal value which is 1 if the literal is positive and 0 if negative.

*Equation 5.*

$$proximity(l, c, v) = \begin{cases} 1 - v & \text{if } l \text{ is a positive literal in } c \\ v & \text{else} \end{cases} \tag{5}$$

When all the activation values *y* corresponding to the clause variables are opposing their desired values, the clause is violated and the product of the distances is exactly 1. If at least one of the variables has an activation that is exactly the desired value, the clause is satisfied and the product of the distances is 0. When the activation values are not exactly Boolean, the *ProP* function outputs a real number in *(0,1)*. For example, consider a clause *c=(A ∨ B ∨ ¬C ∨ ¬D)* and activation array *y=[0.1,0.2,0.6,0.7]* corresponding to units *A,B,C,D,* then

$$ProP(c,y)=(1-0.1)(1-0.2)(0.6)(0.7) = 0.3024$$

The *ProP Vloss* of a multi-clause CNF is the (weighted) average of the *ProP ClauseLosses* per clause (Equations 3 and 4) and can be brought into a sum-of-weighted-products form. For example, assuming penalties of 1, the *ProP Vloss* of the 2-clause CNF, *(A ∨ B ∨ ¬C ∨ ¬D) ∧ (¬C ∨ D),* is the average of the two *ProP ClauseLoss* functions: $\frac{1}{2}$ *((1-* $A)(1-B)CD+C(1-D))= \frac{1}{2}$ $((CD-ACD-BCD+ABCD)+(C-CD))$ which may be re-written as the weighted sum of products: *0.5C- 0.5ACD-0.5BCD+0.5ABCD*.

The partial derivative of *the ProP ClauseLoss* follows (Equation 6):

*Equation 6.*

$$\frac{\partial ProP(c,y)}{\partial v} = \begin{cases} -\prod_{l \in c \;\land\; l \neq v} proximity(l,c,y_l) & \text{if } v \text{ is positive literal in } c \\ \prod_{l \in c \;\land\; l \neq v} proximity(l,c,y_l) & \text{else} \end{cases} \quad (6)$$

For example, given $c = (A \lor B \lor \neg C \lor \neg D)$, the partial derivative of the *ProP ClauseLoss* with respect to a positive literal A is:

$$\frac{\partial prop(c,y)}{\partial A} = \frac{\partial (1-A)(1-B)CD}{\partial A} = -(1-B)CD = CD - BCD$$

whereas the partial derivative with respect to a negative literal is:

$$\frac{\partial prop(c,y)}{\partial C} = (1-A)(1-B)D = D - AD - BD + ABD$$

Notice that the partial derivative of the *ProP Vloss* is the average of the partial derivatives per clause and therefore also has the form of a sum of weighted products. A sigma-pi unit can therefore calculate the partial derivative of the *ProP*-loss using multiplicative connections. This ability of direct gradient computation by sigma-pi units enables the *ProP Vloss* function to act as an energy function which is minimized by the symmetric architecture, rather than just as a loss function to guide learning.

Note also that when the chain rule is used (as in backpropagation) in conjunction with sigmoidal-like activation functions, the *ProP ClauseLoss* gradient is multiplied by $v(1-v)$, which is the derivative of the sigmoid. Equation 7 shows the partial derivative of the *ProP ClauseLoss* with respect to *z*.

*Equation 7.*

$$\frac{\partial Pr(c,y)}{\partial z} = \frac{\partial prop(c,y)}{\partial v}\frac{\partial v}{\partial z} = v(1-v)\begin{cases} -\prod_{l \in c \;\land\; l \neq v} prox(l,c,y) & \text{if } l \text{ is positive in } c \\ \prod_{l \in c \;\land\; l \neq v} prox(l,c,y) & \text{else} \end{cases} \quad (7)$$

This may cause gradients to diminish when the units approach their extreme 0 or 1 values. Empirically this phenomenon was hardly observed in the implementation.

**Log-Satisfaction *Vloss*.** In yet another single-clause *ClauseLoss* function, the log of the satisfaction degree is measured using Equation 8. The *LogSat* ClauseLoss returns therefore values between *0* (satisfaction) to *infinity* (violation).

*Equation 8.*

$$logSat(c,y) = -log(max_{l \in c}\{1 - proximity(l,c,y)\}) \quad (8)$$

Intuitively, when one of the literals is in close proximity to its desired value in the clause, the clause is "almost" satisfied. When this happens, the *max* returns a value that approaches 1 and the *"-log"* therefore approaches 0. When the *y* values approach the opposite of their desired values, the clause approaches violation, i.e., the *"max"* returns a value near 0, and the *"-log"* value approaches infinity. The partial derivative of *LogSat(c,y) ClauseLoss* with respect to a variable *v* in *y* is given in Equation 9 (ignoring non-differentiable points).

*Equation 9.*

$$\frac{\partial logSat(c,y)}{\partial v} = \begin{cases} -\frac{1}{v} & \text{if } v \text{ is a positive literal in } c \text{ and has minimal proximity} \\ \frac{1}{1-v} & \text{if } v \text{ is a negative literal in } c \text{ and has minimal proximity} \\ 0 & \text{else } (y \text{ is not minimal}) \end{cases} \quad (9)$$

Equation 10 is a result of using the chain rule for calculating the *LogSat* partial derivative with respect to *z*, assuming sigmoidal-like activations are used. Note that the result of multiplying Equation 9 by *v(1-v)* has no diminishing terms.

*Equation 10.*

$$\frac{\partial logSat(c,y)}{\partial z} = \frac{\partial logSat(c,y)}{\partial v}\frac{\partial v}{\partial z} = \begin{cases} v-1 & \text{if } y \text{ is positive and minimal} \\ v & \text{if } y \text{ is negative and minimal} \\ 0 & \text{else} \end{cases} \quad (10)$$

When using SGD, a step is taken opposite the gradient sign. Therefore, the *update step* becomes proportional to *(1-v)* for a positive literal and *(–v)* for a negative one. This is intuitive, since in a violated clause, the variable closest to its desired value should be incremented if it is a positive literal and decremented otherwise. The update step size should be proportional to the proximity. For example, when *c=(A ∨B ∨¬C ∨¬D)* and *y=[0.1,0.2.0.6.0.7]* the gradient of the *LogSat ClauseLoss* with respect to *z*'s is the array *[0,0, 0.6,0]*, as *C* is the variable with minimal proximity to its desired value, and is a negative literal in *c*. To compute the *Vloss* gradient when the CNF contains several clauses augmented by penalties, the gradients are computed for each clause and then averaged according to the penalties.

### Algorithms implemented for high-order CONSyN architecture

In the CONSyN architecture, the *ProP Vloss* function is used for two purposes: In the first, it is used as a base for a high-order energy function that controls the dynamics of the network and is gradually modified by the learning process. In the second, the *ProP Vloss* is used to guide the learning process, thus shaping the energy surface in order to find solutions faster. Although several symmetric ANN paradigms could be used (e.g., RBM, Belief Networks, MFT), a simple high-order Hopfield-style network was implemented with no hidden units at all.

When the units of the CONSyN are activated asynchronously, the network may be viewed as minimizing a high-order energy function using SGD. The units reverse their activations in a direction opposite to the direction of the energy gradient which is computed by each unit by summing weighted input products (*z*). Learning in this paradigm means that small changes are made to the energy surface when the network settles on a violating local minimum, so that the energy of violating units is "lifted" while the energy of non-violating units remains unchanged

### *ProP Vloss* as an energy function

The energy function that is being learned is specified in Equation 11. It is very similar to the *ProP* loss function; however, there is no need to average and the penalties ($\beta$) are learned and need not be identical to the pre-set penalties ($\alpha$) of the loss function.

*Equation 11.*

$$E_{CNF}(y) = \sum_{c \in CNF} \beta_c ProP(c,y) = \sum_{c \in CNF} \beta_c \prod_{l \in c} proximity(l,c,y) \qquad (11)$$

As shown for the *ProP Vloss*, this energy function and its gradient can be rewritten as the sum of weighted product terms and therefore could be directly implemented as a symmetric high-order network that minimizes the energy using SGD. The minimizing network consists of visible Hopfield-like sigma-pi units that correspond to the variables *y* of the energy, symmetric high-order connections that correspond to the product terms of the energy and connection weights which are the coefficients of the products terms. This network can be generated automatically from the CNF with penalties that are pre-set ($\beta=\alpha$). For example, the network of Figure 10 is a result of pre-assigning a penalty of $\beta=1$ for the clauses in the CNF $(A \vee B \vee \neg C) \wedge (\neg A \vee \neg B \vee \neg C)$. The energy minimized by that network is $E(A,B,C) = (1-A)(1-B)C+ABC = C-BC-AC+2ABC$, while the network's weights are the coefficients of the energy terms (reversing the sign).

When the $\beta$-penalties are all positive, Equation 11 is globally minimized in exactly the satisfying solutions of the CNF and when the CNF could not be satisfied, the global minima are those solutions with minimum violation (sum of $\beta$'s of violated clauses).

As an example, the network shown in Figure 11 searches for satisfiability of a CNF consisting of clauses $c_1 = (A \vee B \vee \neg c)$ and $c_2 = (\neg A \vee B))$ with assigned penalties $\beta_1$ and $\beta_2$. The network's energy is $\beta_1 ProP(c_1,y) + \beta_2 ProP(c_2,y)$ where $y=[A,B,C]$ is the array of CNF variables. This network was generated by computing the *ProP ClauseLoss* function for each clause:

$$ProP(c_1,y) = (1-A)(1-B)C = C-AC-BC+ABC$$

$$ProP(c_2,y) = A(1-B) = A-AB$$

Summing the penalized terms: $E_{CNF}(y) = \beta_1 C - \beta_1 AC - \beta_1 BC + \beta_1 ABC + \beta_2 A - \beta_2 AB$

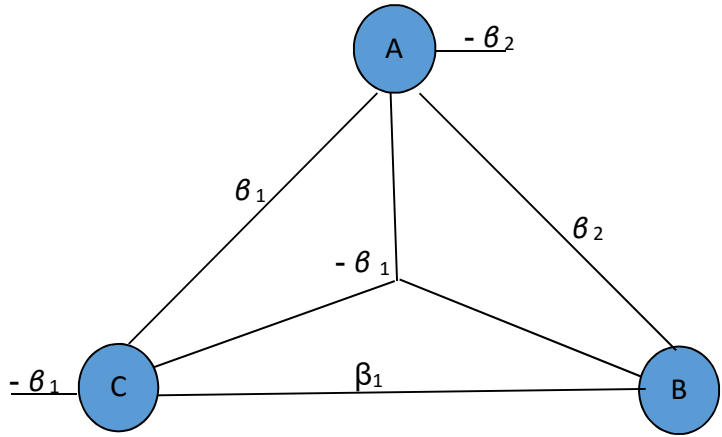

*Figure 11:* A high-order symmetric network that searches for a satisfying solution for *(A ∨ B ∨ ¬C) ∧ (¬A ∨ B)* with corresponding learnable penalties $\beta_1$ and $\beta_2$.

The units of Figure 11 compute (asynchronously) the energy gradient and reverse their activations away from the gradient direction, thus performing SGD down the energy. In the experiments of the next chapter, an initial network has been generated in two varieties: The first is a "compiled" network generated using connections derived from the *ProP Vloss* function with pre-set weights $\beta = \alpha$. The second is a "random" weight network where only the connection products are derived from the *Prop Vloss,* while the weights are randomly initialized in [-1,1].

**Searching for satisfying solutions in an energy minimization network**

As seen, by assigning positive penalties ($\beta$) in the energy of Equation 11, the generated CONSyN network searches for solutions with minimal violation, yet it may be plagued by local minima that only partially satisfy the CNF. The task of learning, therefore, is to adjust the connection weights to avoid local minima. The hope is that by using this kind of energy re-shaping, some local minima will disappear and the average time required to find a satisfying solution will be reduced. Although the learning process changes the energy surface, the $\beta$-penalties remain positives, and it is therefore guaranteed that the satisfying solutions of the CNF (if any) are equal to the global minima of the energy. The following is a specific implementation of Algorithm 1 for solving a problem instance using a CONSyN network with Hopfield-like activations:

*Algorithm 3.* Implemented CONSyN constraint solver

Given:
- Augmented CNF- shared constraints augmented with penalties
- Inputs for clamping- a set of literals to be clamped for a particular problem instance
- CONSyN- with symmetric connections supporting the *ProP Vloss* product terms

Given hyper-parameters:
- *SelectedVClauses*- specifying the number of violated clauses to learn from
- *WeightBound*- specifying a bound for the maximal connection weight allowed
- *MaxSoft*- specifying the maximal number of soft constraint violations allowed

a. Clamp Inputs onto the input units.
b. Set random Binary activation values to all non-clamped units.
c. Calculate *Activation until Convergence* (Algorithm 4).
d. While (some hard constraints or more than *MaxSoft* soft constraints) are violated, do *Learn-Activate* iterations:
  i. CONSyN learning:
     Randomly select violated clauses $c$ (up to *SelectedVClauses* such clauses)
     For each selected $c$, for each connection $s$ containing the negative literals of $c$,
       do *Generalized Anti-Hebb* Learning:
         - If the number of zero units in $s$ is even, decrement weight $w_s$
           else increment $w_s$
         (weight change is scaled so that at least one clause unit flips value).
         - If $|w_s| >$ *WeightBound*, downscale all weights by a factor of 0.01.
  ii. Calculate *Activation until Convergence.*

**The activation convergence calculation process**

In the activation step, the network sigma-pi units are asynchronously activated using the binary threshold function of Equation 12, until an energy minimum is obtained. Algorithm 4 describes the way activation convergence calculation was implemented.

*Equation 12.*

$$y = \begin{cases} 1 & if\ z > 0 \\ 0 & if\ z < 0 \\ Flip & else \end{cases} \quad (12)$$

*Algorithm 4. Activation until Convergence* calculation - Hopfield–like

Given:

- Symmetric network (CONSyN) and unit activations array

- *MaxRandmFlips*: the maximal number of consecutive random flips (hyper-parameter)

Repeat loop

a. *Must flips:* While there exists an unstable unit $i$; i.e. ($y_i=1 \wedge z_i<0$) or ($y_i=0 \wedge z_i>0$), randomly select such unit $i$ and flip its value.
b. *Random Flips*: if there exist unit $i$ with $z_i =0$, then randomly select such unit $i$ and flip its value.
c. If *MaxRandomFlips* consecutive random flips were done, exit loop.

The activation values $y$ at the end of the loop are considered the local minimum of the energy function, despite the fact that the random walk in a plateau (Step b) may have been terminated too early (by Step c).

**A generalized anti-Hebb learning rule (CONSyN)**
In SGD, when clauses are violated, the weights of the network should change in a direction opposite to the gradient of the *ProP Vloss*. This update step can be simplified into a Hebbian rule that is generalized to high-order connections (see Equation 16).

Intuitively, activations that violate constraints should be "unlearned," so that the energy associated with such activations is increased. This is done by weakening "supporting" connections and strengthening "unsupporting" connections in a process that reminds the "sleep" phase in "wake-sleep" algorithms such as in Hinton, Osindero & Teh, (2006). In the following, we provide mathematical justification for this generalized anti-Hebb rule.

*Equation 13.*

$$\Delta \beta_c = -\lambda \frac{\partial PropVloss(cnf, y)}{\partial \beta_c} = -\lambda \sum_v \alpha_c \frac{\partial ProP(c,y)}{\partial v} \frac{\partial v}{\partial z_v} \frac{\partial z_v}{\partial \beta_c} \quad (13)$$

The only learnable parameters in the energy of Equation 11 are the $\beta_c$'s which "weigh" each clause. Therefore, when performing SGD down *ProP Vloss*, the update rule for $\beta_c$ is obtained (Equation 13) using the chain rule, where $\lambda$ is the learning rate and $\alpha_c$ is the pre-set penalty of the augmented clause $c$. Notice however, that in the symmetric paradigm, each unit $v$ computes $(z_v)$ which is (minus) the partial derivative of the energy function as in Equation 14(a).

Therefore, when computing the partial derivative of *z* with respect to $β_c$, Equation 14(b) is obtained. Combining Equations 13 and 14(b) results in Equation 15.

*Equation 14.*

(a)            (b)

$$z_v = -\sum_c β_c \frac{\partial ProP(c,y)}{\partial v} \qquad \frac{\partial z_v}{\partial β_c} = -\frac{\partial ProP(c,y)}{\partial v} \qquad (14)$$

*Equation 15.*

$$\Delta β_c = \lambda \sum_v α_c β_c \left( \frac{\partial ProP(c,y)}{\partial v} \right)^2 \frac{\partial v}{\partial z} \qquad (15)$$

From Equation 15 and knowing that $\frac{\partial v}{\partial z} \geq 0$, it follows that the *β*-weight of a violated clause should always increase while executing SGD down the *ProP Vloss*. Although somewhat surprising, this result is quite intuitive; whenever the network falls into a local minimum that violates a clause, the learning process changes the connection weights in order to strengthen the *β*-weight for the violated constraints and therefore "lifts" the energy of the violating network state.

    A single clause may affect many connections, thus incrementing the penalty is not a local weight change as one would wish for a neural network learning. Luckily, from observing the *ProP ClauseLoss* function, it is possible to deduce how each of the weights involved will change when increasing the *β*-penalty as follows:

    Looking at the *ProP Vloss* of Equations 4 and 5, one can observe that the sum of products, that is the result of multiplying the proximities, consists of positive and negative product terms. Each relevant product term includes the negative literals of the clause, while its sign depends on the number of positive literals in the product. The sign is positive when the product involves an even number of positive literals and is negative when an odd number of positive literals are involved. For example, for Clause c1: *(A ∨ B ∨ ¬C)*, the *ProP* product terms are: *+C-AC-BC+ABC*, and all include the negative literal *C*. Product terms with an even number of positive literals (such as *ABC* or *C*) are positive, and those with an odd number of positive literals (such as *–AC-BC*) are negative terms. When these energy terms are translated into connections, the signs are reversed for SGD. In the example, the weight of the 3rd order connection *ABC*, should decrease because it includes an even number of positive literals *(A,B)*. Similarly, the weight of the pairwise connection *BC* increases because it consists of an odd number of positive literals *(B)*.

    Fortunately, the parity of the positive literals can be sensed from the activation of the units when the clause is violated. Upon violation of the clause, the unit values corresponding to the positive literals are 0s (representing Boolean *"false"*) while those corresponding to the negative literals are 1s (representing *"true"*). It is therefore only necessary to compute the parity of the *false* units. If the number of *false* units in a connection related to a violated clause is even, the weight should decrease, otherwise, the weight should increase (Algorithm 3-*i*). This parity

rule may be put elegantly when the Boolean values of the units have bipolar representation (1 for *true*, -1 for *false*) as in Equation 16:

*Equation 16. Anti-Hebb rule for symmetric high-order connection*

$$\Delta w_s = -\lambda \prod_{y_j \in S} y_j \tag{16}$$

where *S* is a symmetric connection of bi-polar units. When clause *c* is violated, a relevant connection includes all the negative literals which are all ones, whereas the positive literals are all *-1*. By multiplying the unit values, the parity of the number of *-1s* is calculated.

Intuitively, the anti-Hebb rule means that whenever a clause *c* is violated, its related connections are "unlearned" in the following way: if the number of *false* units in a connection is even, they "excite" each other and therefore, their weight should decrease. Otherwise, when the number of *false* units is odd, they "inhibit" each other. Thus, their weight should be increased.

To see that this is an extension of the familiar anti-Hebb rule, consider a pairwise synapse. If the two units "fire" together (or "silent" together), then they should be "unwired," and the connection weight should weaken. If one of the neurons fires and the other is silent, there is an odd number of false units, and the "wiring" between them should be strengthened.

Note that in implementing Algorithm 3, the learning rate $\lambda$ is determined automatically per each clause that is violated; i.e., for each violated clause, $\lambda$ is set to be the minimal that would cause at least one unit within the violated clause to flip its value.

### Algorithms for high-order CONSRNN architecture

As seen in Figure 7, the RNN architecture consists of a feed forward network for mapping the input layer into the output layer with a feedback loop. Although hidden units and deep architecture may be used, high-order connections are implemented instead, with no hidden layers at all. Thus, in our implementation, the output layer is then copied (with some noise) back into the input layer. After each feed-forward step, if the output still violates the constraints, learning is done by computing the gradient of either the *LogSat Vloss* or the *ProP Vloss*. Algorithm 5 is the more detailed implementation of Algorithm 2, using sigma-pi output units and no hidden layers. The connections that are used in the implementations are those specified by the product terms of the *ProP Vloss* with the addition of full pairwise connectivity (input layer to output layer) and biases in the output layer.

*Algorithm 5.* Implemented CONSRNN constraint solver

   Given:
      - Augmented CNF
      - Inputs for clamping
      - CONSRN network - with connections (at least) supporting the *ProP Vloss* product terms
   Given hyper-parameters:
      - *MiniBatch*: the number of learnings iterations before weight update is made
      - *NoiseLevel*: probability of randomizing a non-clamped input unit after each iteration
      - *NoImprove*: the number of non-improving iterations before reinitializing inputs
      - *MaxSoft*: the maximal number of soft constraints allowed in a solution

a. Clamp inputs.
b. Set random (0-1) initial activation values to the non-clamped input units.
c. *Feed-forward computation*: calculate activations for the output layer.
d. While (some hard constraints or more than *MaxSoft* soft constraints) are violated, do (*Learn-Activate iteration*):
   a. *CONSRNN learning*: weight changes using the *noisy-δ-rule* (Algorithm 6),
   b. Every *MiniBatch* iterations, average the changes and update the weights.
   c. If there is no improvement in violation for *NoImprove* iterations, restart by assigning random (0-1) values to the non-clamped input units;
   d. else, copy the output layer onto the non-clamped inputs, while randomizing (with probability *NoiseLevel*) the values of the non-clamped inputs.
   e. *Feed forward computation*,

Learning, as typically done in backpropagation, involves computing the gradient of the *Vloss* with respect to the weights, using the chain rule as in Equation 17, where $S_v$ is a directed connection from a set $S$ of input units to output unit $v$.

*Equation 17:*
$$\Delta w_{s,v} = -\lambda \frac{\partial Vloss(cnf, y)}{\partial w_{s,v}} = -\lambda \frac{\partial Vloss(cnf, y)}{\partial z_v} \frac{\partial z_v}{\partial w_{s.v}} = \lambda\, \delta_v \frac{\partial z_v}{\partial w_{s.v}} \quad (17)$$

The partial derivative of the -*Vloss(cnf,y)* function with respect to $z_v$, is the error ($\delta_v$) per unit $v$. Equation 18 provides the error for unit $v$ as a weighted average of the errors per clause:

*Equation 18. The δ-error for unit v*
$$\delta_v = -\frac{\partial Vloss(cnf, y)}{\partial z_v} = -\frac{1}{\sum_c \alpha_c} \sum_c \alpha_c \frac{\partial ClauseLoss(c, y)}{\partial z_v} \quad (18)$$

Of course, the error per clause depends on the gradient of the specific *ClauseLoss* type, which is either *ProP* (Equation 7) or *LogSat* (Equation 10).

Algorithm 6 calculates $\delta_v$ error for each of the output units with the addition of noise. Empirically (see next chapter), adding noise to the error calculation process is found to improve the generalization abilities of the CONSERN architecture.

Calculating the *noisy-error* requires a pass over all clauses. For each clause, and each unit within the clause, the partial derivative of the *Vloss* with respect to the unit value is calculated. However, with some pre-specified probability, instead of using the *ClauseLoss* derivative (as in Equations 7 or 9), one unit is selected randomly from the clause, and the error is calculated only with respect to that variable.

*Algorithm 6.* Noisy δ calculation

Given a CNF and an activation array *y*
Given hyper-parameters:

- *NoisyGradProb*: The probability of selecting a clause for random error calculation
- *LearningRate*: a positive real
- *Mode=PropVloss/LogSatVloss*: The specific type of *Vloss* function to be used

1. For each violated clause $c$ in the CNF calculate the clause error:
   a. With *NoisyGradProb* probability, select a random unit $v$ from clause $c$ and compute its error to be *1-v* if $v$ is positive in $c$ and *–v* otherwise.
   b. Else (*1-NoiseyGradProb* probability),
      If *Mode=LogSatVloss*, select the variable with minimal proximity among the clause variable. Calculate its -error using Equation 10;
      else, (*Mode=PropVloss*) for each unit $v$ in the clause:
      calculate its -error using Equation 7.
2. The total error $\delta_v$ is the weighted average of the clause-errors for $v$ (Equation 18).

The update rule for a weight is derived the usual way (from Equation 17) and is provided in Equation 19.

*Equation 19. The delta rule for a high-order connection*

$$\Delta w_{s,v} = \lambda\, \delta_v\, \frac{\partial z_v}{\partial w_{s,v}} = \lambda\, \delta_v \prod_{y_j \in S} y_j \tag{19}$$

## Experimental Results

Both proposed architectures share the same experimental framework: A weighted CNF was generated from the block planning constraint specifications (Table 1) allowing up to six blocks of up to three colors and three sizes and a maximum of seven step plans. The CNF uses 385 variables and 5,832 clauses. Hard clauses were augmented with *α=1000* penalties, and soft clauses with *α=1* penalties. Following the CNF generation, a network (either CONSRNN or CONSyN) was generated with visible units corresponding to the CNF variables and connections derived from the *ProP(CNF,y) Vloss* function. The generated network was given a training set of planning instances to solve. Each training instance was solved by the network, and the weights learned during the solving were carried to the next training instance. Every 10 training instances, the network was tested by solving a set of 50 test instances.

The weights of the CONSRNN were randomly initialized, while in the CONSyN architecture, a comparison was made between the performance of random weight initialization and the performance of compiled weights initialization (*β= α*). Algorithm 3 (for CONSyN) and Algorithm 5 (for CONSRNN) were used for both training and testing instances. However, testing, the weights were not transferred from one instance to the next.

### Generation of Train and Test Planning Instances

Random block-world instances were generated with three colors and three sizes. In total, 450 instances were generated in three different levels of difficulty: 150 3-block problems (*easy*), 150 4-block problems (*medium*), and 150 5-block problems (*difficult*). Each problem instance included a randomized assignment of *color* and *size* for the blocks, a random initial block arrangement, and a random goal arrangement. For each level of difficulty, 100 instances were used as a training set, and 50 different instances were used as a test set. A set of 50 3-block instances was used as validation to select some of the hyper-parameters.

As a proxy for performance, for each test instance, the number of steps required to obtain a satisfying solution was measured. In the CONSyN architecture, the number of unit flips was measured, whereas in the CONSRNN case, the number of activate-learn iterations was measured.

In each experiment, the test set was tried prior to training (point *0* in all graphs) and then again every 10 training instances. Each experiment was repeated 10 times, each time with a different (randomized) order of the training instances. The performance measured was averaged across the 50 test instances and then across 10 different experiments (with different randomization).

**Results for the CONSyN Architecture**

The experiments of the symmetric network were conducted using Algorithm 3 and 4 running on a symmetric network generated with connections that correspond to the *ProP Vloss* terms of the CNF. Performance of a test instance was measured by the number of unit flips that were made until a solution was found. Figure 12 shows the average performance of a compiled CONSyN trained using training sets of various difficulty levels. Hyper-parameters used: *SelectedVClauses*=1, *WeightBound*= 200,000, and *MaxSoft*=100.

The C3x3 graph shows the test performance prior and during a training session of 100 3-block instances. The untrained (compiled) network starts at an average of about 1,400 flips per instance. However, after 10 training instances, the network gains speed (average flips is 385) and after just 20 training instances, the average number of flips for solving a 3-block test problem is stabilized around 250 flips.

The rest of the graphs in Figure 12 are tested on more difficult (5-block) instances. The G5x5, G4x5, and G3x5 graphs show the result of a 5-block test performance while practicing on 5, 4, and 3 blocks respectively. Not surprisingly, training on difficult (5-block) problems gives the best result. Surprisingly, the network is capable of generalizing from practicing on easy problems when tested on the more difficult 5-block training set. Training using four blocks speeds-up better than training on three blocks. However, training first on 20 easy (3-block) instances and then on the more difficult four blocks achieved the performance of 5-block training after about 60 trainings (see G34x5 graph). However, when continued with 4-block trainings, the test performance degraded and approached the speed of the G4x5 graph.

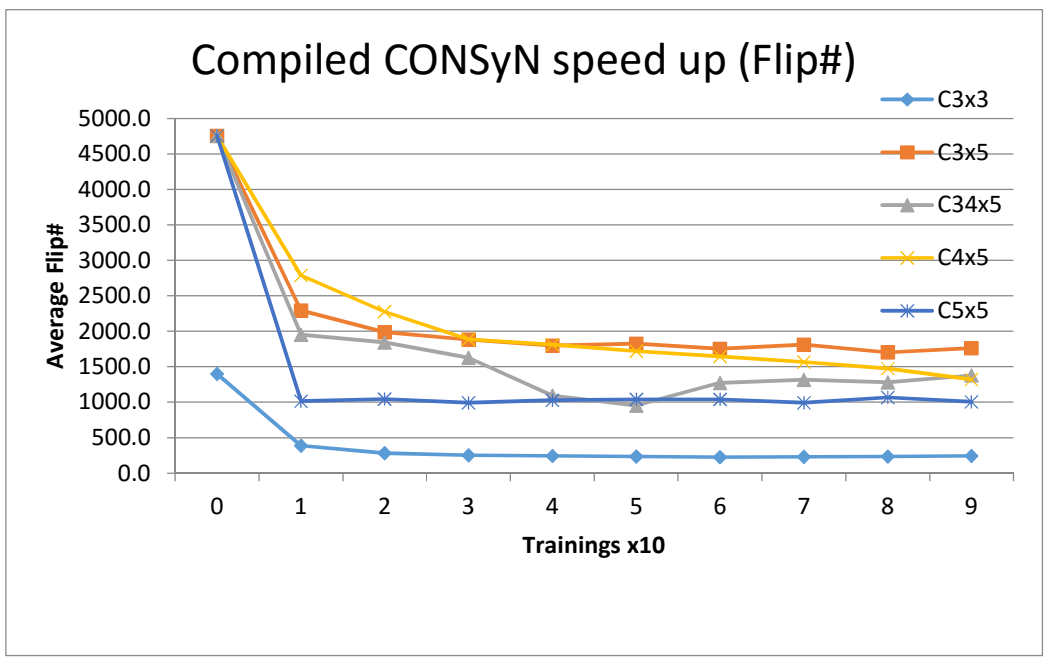

*Figure 12.* Performance (in flips) starting with compiled weights

In Figure 13, the performance of compiled networks (COMP) on 5-block test problems is compared with randomly initiated networks (Rand). The random networks are extremely slow just prior to training, but after few training instances, performance is accelerated and approaches that of the compiled network.

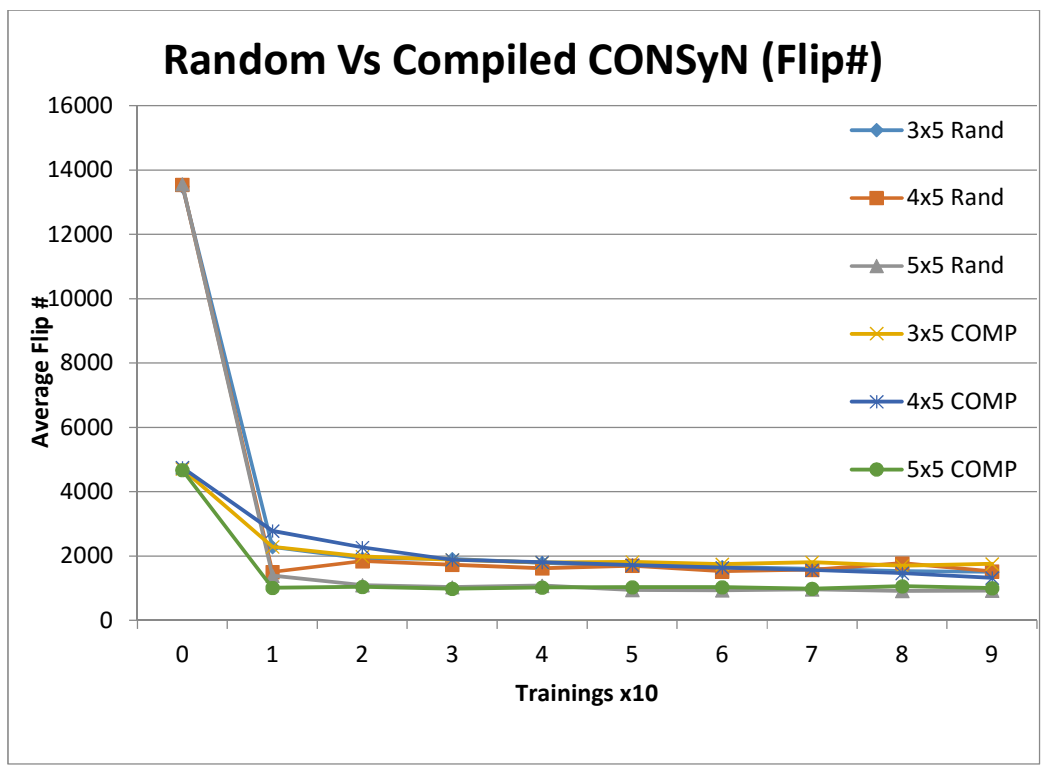

*Figure 13.* Comparison of randomly initiated networks with compiled networks (5X5)

In Figure 14, the performance of a compiled CONSyN network of 5-block training and 5-block testing was measured using the average number of "activation-learn" iterations instead of the number of flips. The shapes of the graphs measuring iterations seem to resemble those measuring flips. Prior to the training, it took an average of 170 iterations to solve a 5-block test instance and after just 20 training instances, the number of trainings needed dropped to about 50 iterations on average. This may signal that the number of interfering local minima drops.

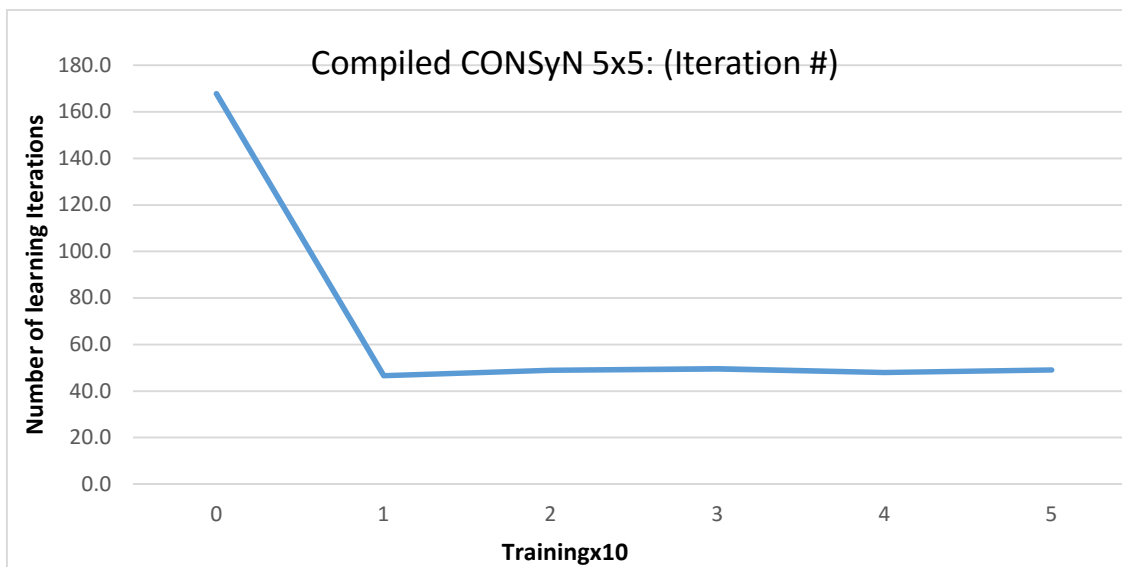

*Figure 14.* Compiled CONSyN network (5x5) measured using the number of iterations

**Results for the CONSRNN Architecture**

The experiments of the CONSRNN were conducted using Algorithm 5 running on an RNN with high-order connections and no hidden units. The connections generated included all pairwise connections between the input and output layers, as well as all higher-order *directed* connections that corresponded to all permutations of the *ProP-loss* product terms. For example, if *ABC* is a term in the *ProP Vloss*, then three directed 3-order connections are generated corresponding to the permutations of this particular term: $\{A,B\}_C$ ; $\{A,C\}_B$; $\{B,C\}_A$. The weights were initialized with uniform random real values in (-1,1).

The performance of the CONSRNN was tested using the same methodology. However, the number of "activate-learn" iterations was measured instead of the number of flips. As in the CONSyN case, each experiment was replicated 10 times with different randomizations, and the performance measured was averaged across the test instances.

Figure 15 shows the average number of iterations needed for solving 3-block instances while training on 3-block instances. The graph compares the result of using the *ProP Vloss vs the LogSat Vloss* and shows similar speedups. In these experiments hyper parameters were selected using the 3-block validation set: *NoiseLevel=0.15, NoisyGradProb=0.06, LearningRate= 0.06*. Similarly, Figure 16 shows the performance speed-up when testing and training 4-block problems, while Figure 17 shows testing and training of 5-block problems.

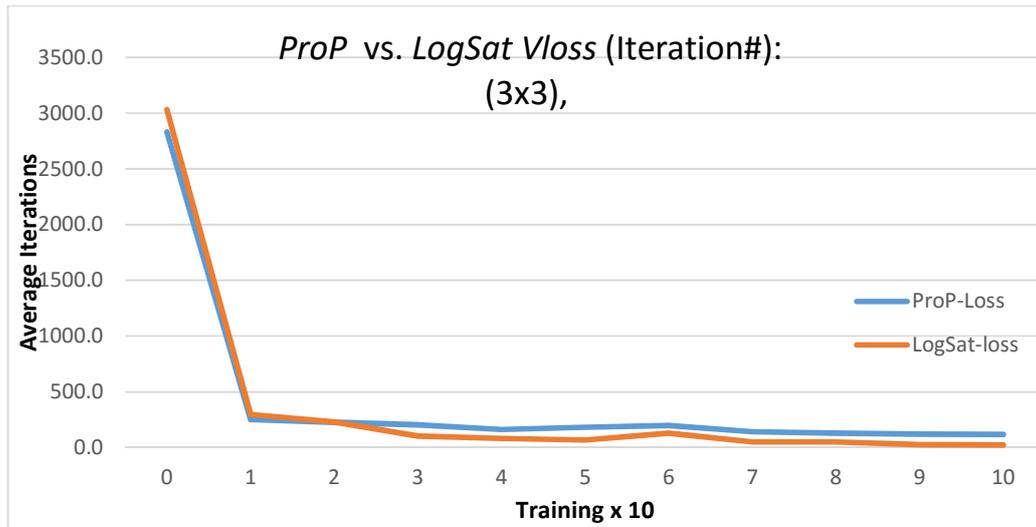

*Figure 15.* Comparing performance of *ProP* vs. *LogSat Vloss* (3X3)

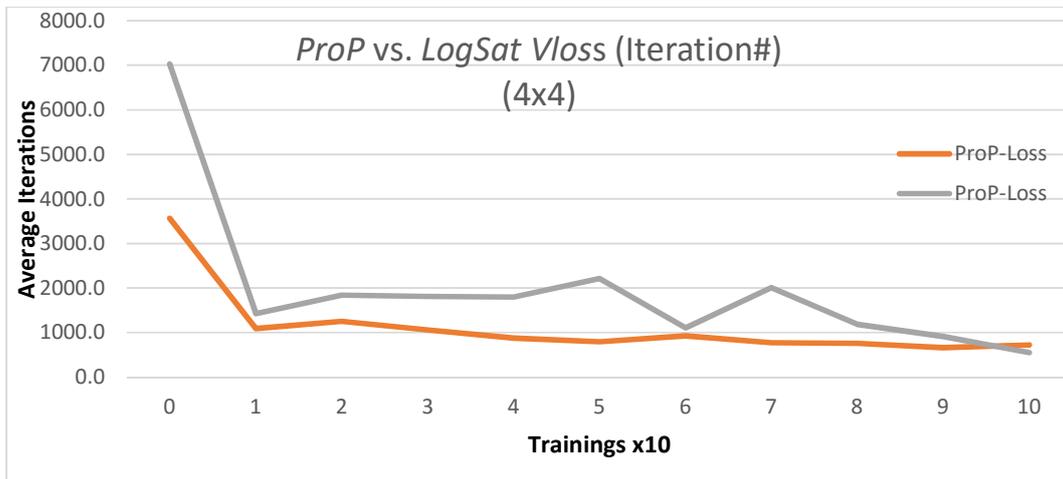

*Figure 16.* Comparing performance of *ProP-loss* vs. *LogSat Vloss* (4X4)

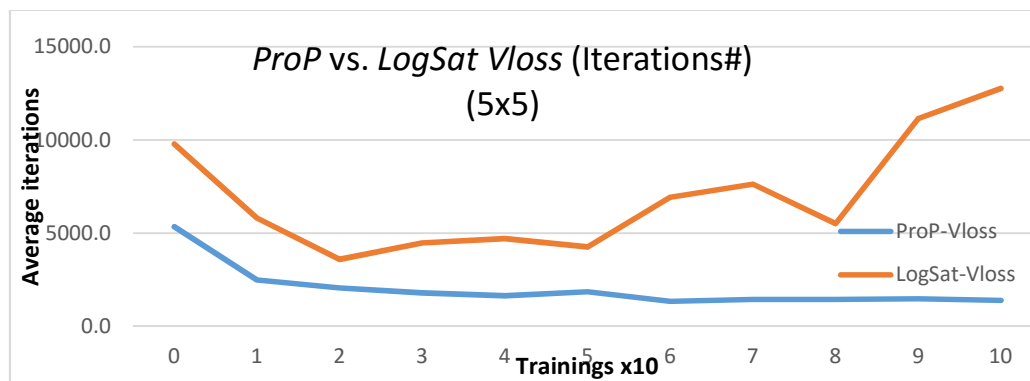

*Figure 17.* Comparing performance of *ProP* vs. *LogSat Vloss (5X5)*

Empirically, on this set of experiments, the *ProP Vloss* provided better performance than the *logSat Vloss* and was less prone to over-fitting. It should be noted however, that the *ProP Vloss* gradient calculation is somewhat more demanding computationally.

In Figure 18, the effect of various noisy gradient probabilities on generalization after just 10 training instances are shown on 3-block training and testing.

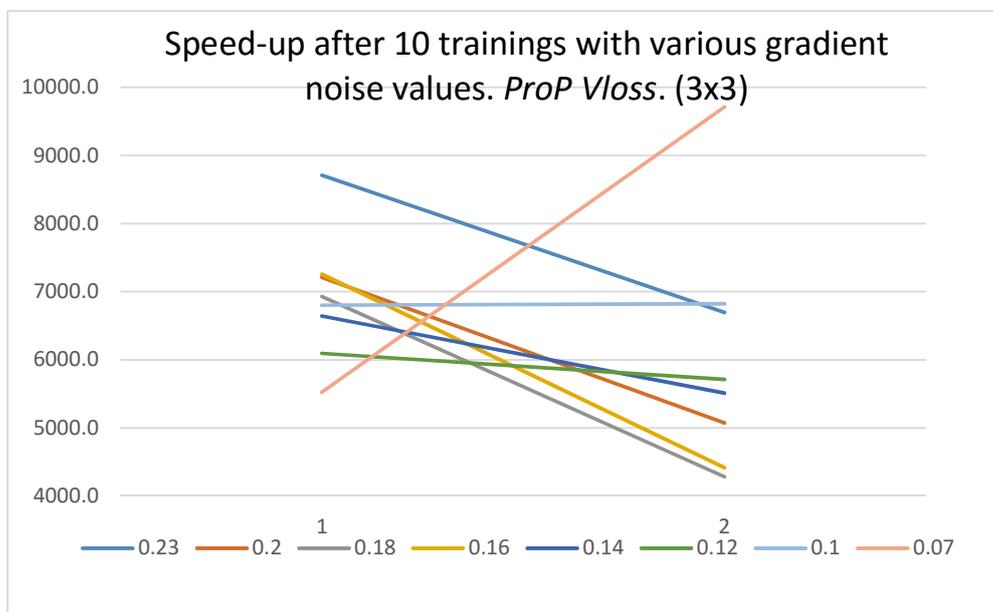

*Figure 18.* Speed-up of *ProP Vloss* (3X5) with various noisy gradient probabilities. Performance was measured before and after practicing 10 training instances.

Noisy gradient probabilities between *0.07* and *0.23* were tried; when the gradient noise level was too small *(<0.12)*, practicing was ineffective and even worsened the performance (0.07). Best performance and generalization were observed with noisy gradient probabilities of *0.16-0.18*. Higher noise levels (e.g., *NoisyGradProb=0.23*) seemed to be able to speed up performance but also led to overall inferior solving abilities.

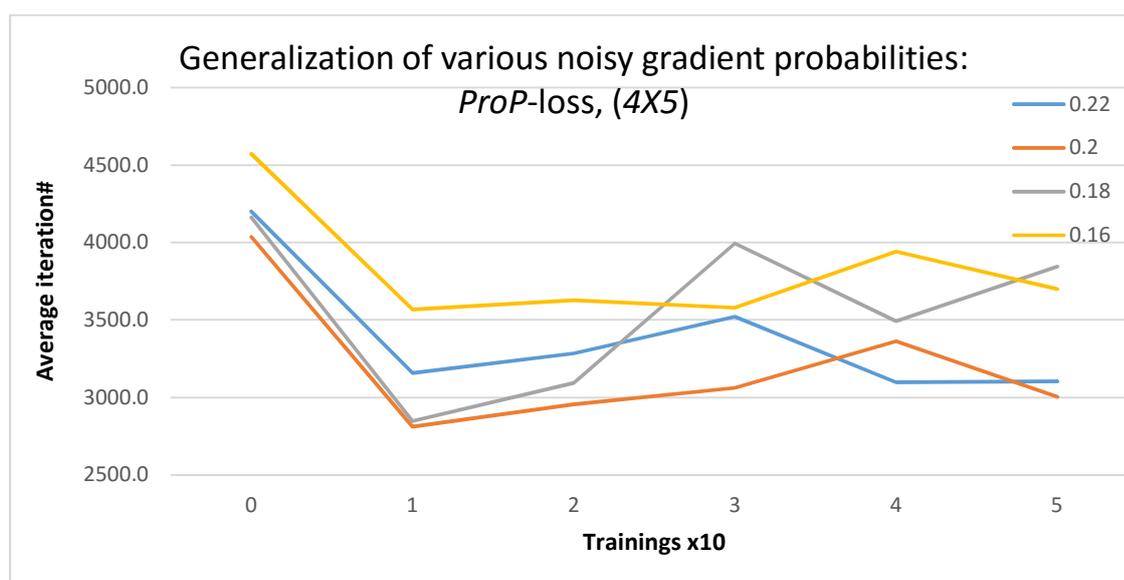

*Figure 19.* Effect of various noisy gradient probabilities on generalization (5-block testing, 4-blocks training)

In Figure 19, the effect of various noisy gradient probabilities is shown on a 5-block test set, while training used 4-block instances. Significant speed-up was observed immediately after the first 10 training instances. However, the test performance deteriorated when training continued as if the network was over-learning. With further training though, this deterioration was corrected in some experiments.

## Discussion

### The Binding Problem

From a cognitive science perspective, some of the time-long challenges of connectionism have been targets of the present work. One such problem is how compositionality is obtained in connectionist networks, i.e., how simple representations, encoded in activation, can be combined and be re-used for forming novel and useful representations (Feldmann, 2013; Fodor & Philishine, 1988). Another challenge is how rules are consistently and systematically used in ANNs (Marcus, 2001). These challenges and others are all related to the long-standing "binding problem" (Feldmann, 2013; Malsberg, 1992; Velik, 2010; Zimmer, Mecklinger & Lindenberger, 2006; Anadan, Levtovsky & Mjolsness, 1989). Exactly how binding can occur in a massively parallel network of simple processors is still a fundamental question in cognitive science, while its variable-binding aspect has been identified as a "nastier" problem and a key to neural theories of language (Jackendoff, 2002).

A (variable) binding mechanism for planning problems has been presented using crossbars that represent relationships among objects and between objects and their attributes. The mechanism presented is general purpose and allows dynamic formation of arbitrary structures made of unit collections (binders) that indirectly refer to simpler constituents. Processing of these structures is done by enforcing learned constraints, so that novel and complex representations emerge on the output units when the constraints are satisfied. Although not directly demonstrated by our planning examples, binders may be dynamically allocated (by constraints) to form arbitrary directed graphs (Pinkas, Lima & Cohen, 2013).

Throughout the years, several neural architectures have been proposed for solving the binding problem (Barret, Feldman & Dermed, 2008; Shastri & Ajjanagad, 2006; Browne & Sun, 2000; Van Der Velde & Kamps, 2006) and implementing logic systems in ANN (Hölldobler & Kurfess, 1992; Stolke, 1989; Ballard, 1986). Nevertheless, the binding mechanism presented here is rather different in the underlying representation principle, expressive power, dynamics, and in its learning ability.

For example, "Shruti" based architectures, both temporal (Shastri & Ajjanagad, 2006), and non-temporal (Barret, Feldman, & Dermed, 2008), compromise expressiveness for the simplicity of "spreading of activation." Only a limited subset of FOL is allowed, but performance is linearly bound by the size of the long-term knowledge represented. In contrast, the binder crossbar mechanism suggested here is highly expressive, and the extra expressive power is obtained by trading time complexity. Training is thus necessary in order to obtain performance speedup while in the "Shruti" based networks, pre-wiring is possible and need not be tuned-up using learning. Another important difference is that in "Shruti", LTM is encoded in the network's structure and ad hoc, carefully crafted circuits must be recruited and pre-wired in order to represent pieces of long-term knowledge. In contrast, in the architectures suggested here, LTM is learned and stored only in connection weights while the network structure is general and independent of the constraints learned. Therefore, new long term knowledge does not impose network structure changes.

The black-box architecture (Van Der Velde & Kamps 2006) seems as expressive as our binder-pool (crossbar) mechanism and similarly can model complex working memory structures in activation only. It need not requite units to represent LTM. However, it needs a pre-defined set of modular circuits and a separate, ad-hoc external controller to integrate the different circuits into useful representations and process them. A simple such controller can be learned (Van Der Velde & Kamps, 2010). However, it is not clear if and how combinatorial search can be learned this way and how FOL rules (with variables) could be learned. Nevertheless, our crossbar-based binding mechanism shares with the black-box architecture the idea that an ensemble of neural units can be used for multiple purposes while acting as a pointer device. This leads in both systems to the capability of representing several instances of the same concept simultaneously each with different attributes - see "problem of one" (Jackendoff, 2002).

**SAT Solvers**

Planning problems have been stated as logic deductions since the early days of artificial intelligence (Fikes & Nilsson, 1971). The idea of reducing planning problems into satisfiability (SAT) was first introduced in (Kautz, McAllester, & Selman, 1996). SAT solvers are used today for a variety of applications from planning to program verification and are considered among the fastest solutions for applied *np*-hard problems (Harmelin, Lifschitz, & Porter, 2008). Some sophisticated SAT solvers learn on the fly while solving a specific problem instance, yet to the best of the authors' knowledge, SAT solvers currently do not carry learned knowledge from one problem instance to another. The ANNs proposed in this paper may be viewed as SAT solvers that learn a particular application domain and adapt to it by transferring the learned weight from one problem instance to the next. In their current state, our simulated ANN implementations may not be a match for the efficient state of the art SAT solvers. However, there may be some hope, that when implemented on the right hardware and when they are given time to practice on a specific domain, such architecture may deliver competitive performance. It is thus interesting to note that the iterative loop in both CONSRNN and CONSyN have some resemblance to the basic loop of *WalkSat* (Kautz, Selman & McAllster, 2004), and the error calculation based on the *ProP Vloss* is reminiscent of the greedy heuristic used in *WalkSat*. In some sense, both architectures perform a strategy similar to the stochastic local search greedy heuristics used by such SAT solvers. Unlike these solvers, though, the heuristic is learned by the ANNs.

**Neuro-Symbolic Systems**

In the area of neuro-symbolic integration (Garcez et al., 2006; Hammer & Hitzler, 2007), emphasis has been given on how symbolic knowledge can be extracted from and injected into ANNs. In Tran and Garcez (2016), augmented conjunctive rules were compiled prior to learning, and this was empirically shown to speed-up learning. However, the rules are limited in their structure and must be hierarchical in their nature. In the architectures proposed, augmented logic expressions with no structure limitations could be injected by either pre-compiling them (as in compiled CONSRNN) or by learning them using a *Vloss* function. Thus, prior knowledge can be injected before, during, and even after doing statistical learning.

Few network architectures have dealt with relational knowledge and FOL (Domingos, 2008; Hölldobler & Kurfess, 1992). However, virtually all of them are based on the Herbrand universe (grounding all terms), which tends to exponentially explode. In contrast, the use of binders as un-grounded pointing objects allows preservation of the original FOL structure without including all grounded terms as propositions. The crossbar pool and the constraints that govern it enable binders to dynamically reference complex structures built on the fly out of basic concepts (Pinkas, Lima & Cohen, 2013).

**Summary**

Two ANN architectures have been introduced that learn to search in a combinatorial search space, which is restricted by a set of logic constraints expressed in bounded FOL. The iterative activate-learn process employed by both architectures may be viewed as a parallel constraint satisfaction, using an adaptive learnable heuristic. The constraints are activated by the inputs (e.g. simultaneously by the initial and goal configurations of a planning instance) and are learned upon failing to satisfy them. After several iterations of activation and learning, a solution emerges on the output units, which satisfies the hard constraints with no more than a pre-specified number of soft constraint violations. The learned constraints may be viewed as long-term knowledge that is stored in connection weights and enforces unit activations to represent a valid solution. The activations of the visible units (working memory) represent dynamically created non-trivial structures consisting of relationships among objects and between objects and attributes. A dynamic binding mechanism is provided based on a pool of binders. Each binder is capable of assuming the role of any object, pointing to attribute values or participating in relations.

The learning process is guided by a loss function that measures the degree of constraint violation of the visible unit activations. Unsupervised learning is used to speed-up network performance by "practicing" on training problem instances. Indeed, significant speed-up is observed in the planning domain after practicing on just 10 training problems when testing on unseen problem instances. Speedup is also observed when training is made on "easy" problems, while testing is done on more "difficult" problems.

Two loss functions have been proposed for measuring the degree of constraint violation. Empirically, a slight advantage has been observed for the *ProP Vloss* over the *LogSat Vloss* in the CONSRNN architecture in some experiments; nevertheless, we still consider both functions as valid implementation options. In the CONSyN architecture, only the *ProP Vloss* was implemented, as it naturally fits the energy minimization paradigm. The *ProP Vloss* is used both to pre-compile the network and to reshape the energy function while learning. The compiled network performs much faster than a randomly initialized network prior to training. However, the random network catches-up rapidly and obtains the performance of the compiled network after just a few training instances. In some experiments, it has been observed that the network slows-down when learning continues past a certain point. This may suggest "over-fitting" and therefore, the use of strategies such as early-termination, drop-out, and regularization.

Notice also that the constraints need not be provided ahead of time as only the *Vloss* gradient is actually needed. Thus an external teacher who pinpoints the error is sufficient and no CNF is needed. For instance, it's enough for the teacher to specify a set of units which violate a clause, so that an error could be calculated for each of those units.

In theory, every search problem that can be reduced to SAT can be implemented in the proposed architectures, yet learning by practicing may only be useful if the problem instances (train and test sets) share many of the constraints, so that the learned constraints could be transferred from one instance to another.

**Future Directions**

There are plenty of technical variations yet to be explored in both architectures. Only a few of them have actually been implemented thus far. Deep learning (Bengio, 2009; Bengio, LeCun & Hinton, 2015) is a promising direction for both architectures. In the implementations presented here, no hidden units were used. However, since high-order connections could be traded with layers of hidden units, deep networks can be used instead or in conjunction with high-order connections. The ample existing research in deep learning could be useful. Convolution (e.g. based on CNF cliques), dropout, regularization, and rectified linear activation may all be relevant. Once hidden layers are introduced, more complex RNNs could be tested such as LSTM (Hochreiter & Schmidhuber, 1997).

The idea of a pool of binders governed by learned constraints has a simple and general underlying computational principle, which can be adapted for a variety of other logic or symbol-driven applications, such as verification, deductive databases, and language processing. Nevertheless, integrating statistical learning with prior knowledge seems to have even greater potential as supervised statistical learning may be executed before, after, or in conjunction with constraint learning. This could be done by combining familiar loss functions (e.g. X-entropy) with *Vloss* functions.

For illustration, consider a network trained with a loss function that integrates *ProP Vloss* with X-entropy classification using a weighted average of the two functions. Such an approach would allow the mixing of symbolic constraint processing and statistical classification at the same time with varying confidence levels on the data and on the constraints. For example, for a visual scene to be analysed, objects and relationships could be classified while at the same time, the classification results should satisfy certain domain and physical world constraints. Both the data and the constraints may be augmented with varying confidence levels. The error, computed by the gradient of such integrated loss function will encapsulate both the statistical error and the violation error. Similarly, in NLP, a network may statistically learn to classify parts of speech. At the same time, however, syntactic and semantic constraints should be satisfied and therefore influence the classification.